\documentclass[10pt,twocolumn,letterpaper]{article}

\usepackage{iccv}
\usepackage{times}
\usepackage{epsfig}
\usepackage{graphicx}
\usepackage{amsmath}
\usepackage{amssymb}

\usepackage{multirow}

\usepackage[pagebackref=true,breaklinks=true,letterpaper=true,colorlinks,bookmarks=false]{hyperref}

\iccvfinalcopy % *** Uncomment this line for the final submission

\def\iccvPaperID{3187} % *** Enter the ICCV Paper ID here

\ificcvfinal\fi

\begin{document}

%%%%%%%%% TITLE
\title{AdaZoom: Adaptive Zoom Network for Multi-Scale Object Detection \\ in Large Scenes}

\author{Jingtao Xu \qquad Yali Li \qquad Shengjin Wang \\
Department of Electronic Engineering, Tsinghua University\\
{\tt\small xjd19@mails.tsinghua.edu.cn {liyali13, wgsgj}@tsinghua.edu.cn}}
%\and

\maketitle
\ificcvfinal\thispagestyle{empty}\fi

%%%%%%%%% ABSTRACT
\begin{abstract}
  Detection in large-scale scenes is a challenging problem due to small objects and extreme scale variation. It is essential to focus on the image regions of small objects. In this paper, we propose a novel Adaptive Zoom (AdaZoom) network as a selective magnifier with flexible shape and focal length to adaptively zoom the focus regions for object detection. Based on policy gradient, we construct a reinforcement learning framework for focus region generation, with the reward formulated by object distributions. The scales and aspect ratios of the generated regions are adaptive to the scales and distribution of objects inside. We apply variable magnification according to the scale of the region for adaptive multi-scale detection. We further propose collaborative training to complementarily promote the performance of AdaZoom and detection network. To validate the effectiveness, we conduct extensive experiments on VisDrone2019, UAVDT and DOTA datasets. The experiments show AdaZoom brings consistent and significant improvement over different detection networks, achieving state-of-the-art performance on these datasets, especially outperforming the existing methods by AP of 4.64\% on VisDrone2019.

\end{abstract}

%%%%%%%%% BODY TEXT
\section{Introduction}

In recent years, significant progress has been achieved in computer vision. Visual object detection has also been extensively studied since it is important in various applications such as  video surveillance and autonomous driving. Existing detectors such as Faster R-CNN~\cite{ren2015faster}, YOLO~\cite{redmon2016you}, and CornerNet~\cite{law2018cornernet} achieve satisfying performance on natural images. However, in practical applications such as Unmanned Aerial Vehicle (UAV) vision, existing detectors perform poorly because the images capture large-scale scenes with wide fields of view and quite small objects.

\begin{figure}[t]
  \centering
  %\subcaptionbox{a}{\includegraphics[width=.5\linewidth]{pic/sketch.eps}}\hfill
%  \subcaptionbox{a}{\includegraphics[width=.5\linewidth]{pic/sketch.eps}}
    \includegraphics[width=1.0\linewidth]{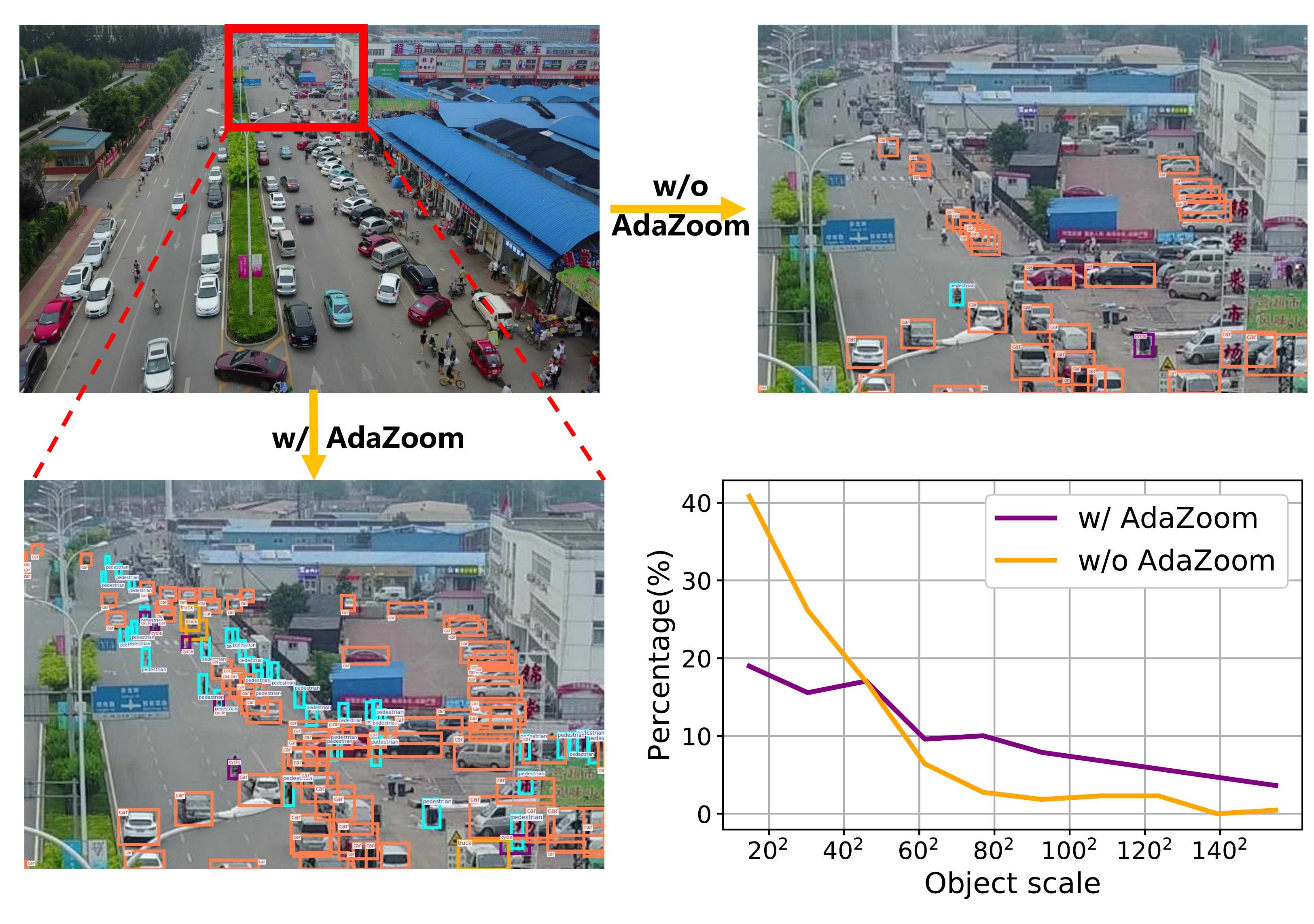}

  \caption{In large-scene images such as UAV images, objects are small and dense. AdaZoom works as a selective magnifier to generate and zoom the focus regions for further detection. The line chart presents the distribution of object scales in the image with and without focus regions generated by AdaZoom, respectively.}

  \label{sketch}
\end{figure}

There are several challenges for detecting objects in large-scale scenes: (1) Objects are small and dense. For example, the UAV takes images from high altitude with wide field of view, as shown in Fig.~\ref{sketch}. Tens and hundreds of objects exist in a single image and most of them occupy quite a few pixels. Deep neural network with successive downsampling would bring intolerable loss for semantic and positional information of small objects, resulting in poor detection performance. (2) Extreme scale variation across objects arises since images of large-scale scenes record a large span of distance and the camera-object distance varies significantly. Objects become smaller with further distance. Even objects of the same category may differ hundred times in scale. However, the receptive field of convolutional neural networks is limited. Extreme scale variation results in semantic gaps in convolution layers and brings substantial burdens in learning the powerful feature representations.

To tackle the object detection in large-scale scenes, it is urgent to design an adaptive zoomer to ``focus'' on objects with varying scales. Although there are some works on resizing images for multi-scale training~\cite{singh2018analysis,singh2018sniper} and inference on enlarged image crops~\cite{gao2018dynamic, najibi2019autofocus, yang2019clustered}, they are inflexible for objects of various scales in large-scale scenes and separate training and inference into different pipelines.

We propose an Adaptive Zoom (AdaZoom) network based on policy gradient~\cite{1999Policy} to adaptively zoom the focus regions for further detection. Inspired by human perception~\cite{mcconkie1975span}, when we perceive a large-scale scene, we glance over the whole image for coarse cognition and zoom where objects are small and dense for a careful watch.

AdaZoom works as a selective magnifier with flexible shape and focal length. It focuses on image regions with small objects. The scale and aspect ratio of the region to be zoomed are adaptive to the scales and distribution of objects inside. For a cluster of smaller objects, AdaZoom prefers a smaller focus region enclosing them for higher magnification, just like using a magnifier with shorter focal length. Without additional annotations for regions, AdaZoom is optimized according to the reward which measures the quality of the focus region. Following the paradigm of deep reinforcement learning, we further learn the policy network to produce focus regions with the regard of visual features. In addition, we design collaborative training to iteratively promote the joint training performance of AdaZoom and the detector. The outputs of detector are introduced to the reward which guides AdaZoom to focus on difficult regions. Then the regions generated by AdaZoom will be zoomed for finetuning the detector. With collaborative training, the detection performance is further improved consistently and significantly.

In summary, the main contributions are as follows:
\begin{itemize}
  \item We propose a novel Adaptive Zoom (AdaZoom) network to adaptively generate and zoom the focus regions for accurate detection in large scenes, without additional annotations for the regions.

  \item We propose collaborative training to jointly boost the coordination between AdaZoom and the detector with a consistent pipeline of training and inference.

  \item Without bells and whistles, AdaZoom achieves state-of-the-art on the VisDrone2019~\cite{zhu2018vision}, UAVDT~\cite{du2018unmanned} and DOTA~\cite{xia2018dota} datasets.
\end{itemize}

%-------------------------------------------------------------------------
\section{Related work}

\textbf{Multi-scale object detection.} Multi-scale training and inference with image pyramid is the most straightforward idea to alleviate the problem of small objects~\cite{dai2016r,2015Fast,2016Deep,2017Object}. However, the image pyramid increases the scale variation of images. SNIP~\cite{singh2018analysis} proposes a training paradigm ignoring objects out of the desire size range during gradient backpropagation. Following the idea of \cite{singh2018analysis}, SNIPER~\cite{singh2018sniper} focuses on efficient multi-scale training by sampling chips of different sizes. These chips are resized to a certain size for multi-scale training. Further, AutoFocus~\cite{najibi2019autofocus} extends SNIPER~\cite{singh2018sniper} to a coarse-to-fine pipeline that predicts regions of interest at a coarse level and then infers on the regions at a fine level. In addition, DREN~\cite{zhang2019fully} and ClusDet~\cite{yang2019clustered} employ neural networks to estimate difficult regions for fine detection. However, these methods separate multi-scale training and inference into different pipelines which leads to inconsistency between training and inference. Apart from multi-scale process in image level, another direction is to build pyramid in feature level~\cite{guo2020augfpn, 2014Spatial,lin2017feature,liu2018path,liu2016ssd}. Feature Pyramid Network~\cite{lin2017feature} is widely utilized in many SOTA detectors for cross-layer feature fusion. Due to the limitation of the receptive field of convolutional neural networks, \cite{cai2016unified, liu2016ssd} assign smaller objects to shallower layers. Our method focuses on image level. We introduce deep reinforcement learning into adaptive region generation for multi-scale object detection and integrate the training and inference into the same pipeline with the regions generated by AdaZoom.

\textbf{Reinforcement-learning-based object detection.} Reinforcement learning is introduced to object detection in the following ways: (1) Focusing on objects step by step with accumulated evidence~\cite{2016Hierarchical,2015Active,gonzalez2015active,hara2017attentional,li2018Object,2016AttentionNet}. In particular, an image contains specific context information and a sequence process accumulates evidence from context for detection. \cite{gonzalez2015active} learns a search strategy to collect context and select the next window to visit. In addition, objects in the same scene have relationship with each other, such as a person riding on a bicycle. The detected objects provide contextual cues for subsequent steps of detection~\cite{kong2017collaborative}. (2) Selecting high-quality regions of interest. \cite{Mathe_2016_CVPR} proposes a sequential exploration process to select region proposals. \cite{pirinen2018deep} introduces deep reinforcement learning into region proposal network to filter out low-quality proposals and accumulate class-specific evidence over time step to boost detection accuracy. Besides, several works~\cite{gao2018dynamic,lu2016adaptive,uzkent2020learning} propose selection strategies to acquire image regions from enormous candidates. Sharing the similar idea, we go beyond in the adaptability of region generation and propose effective collaborative training for region generation and detection.

\section{Method}
\subsection{Analysis of Focus Adaptability}
\begin{figure}[t]
  \centering
  %\subcaptionbox{a}{\includegraphics[width=.5\linewidth]{pic/sketch.eps}}\hfill
%  \subcaptionbox{a}{\includegraphics[width=.5\linewidth]{pic/sketch.eps}}
    \includegraphics[width=1.0\linewidth]{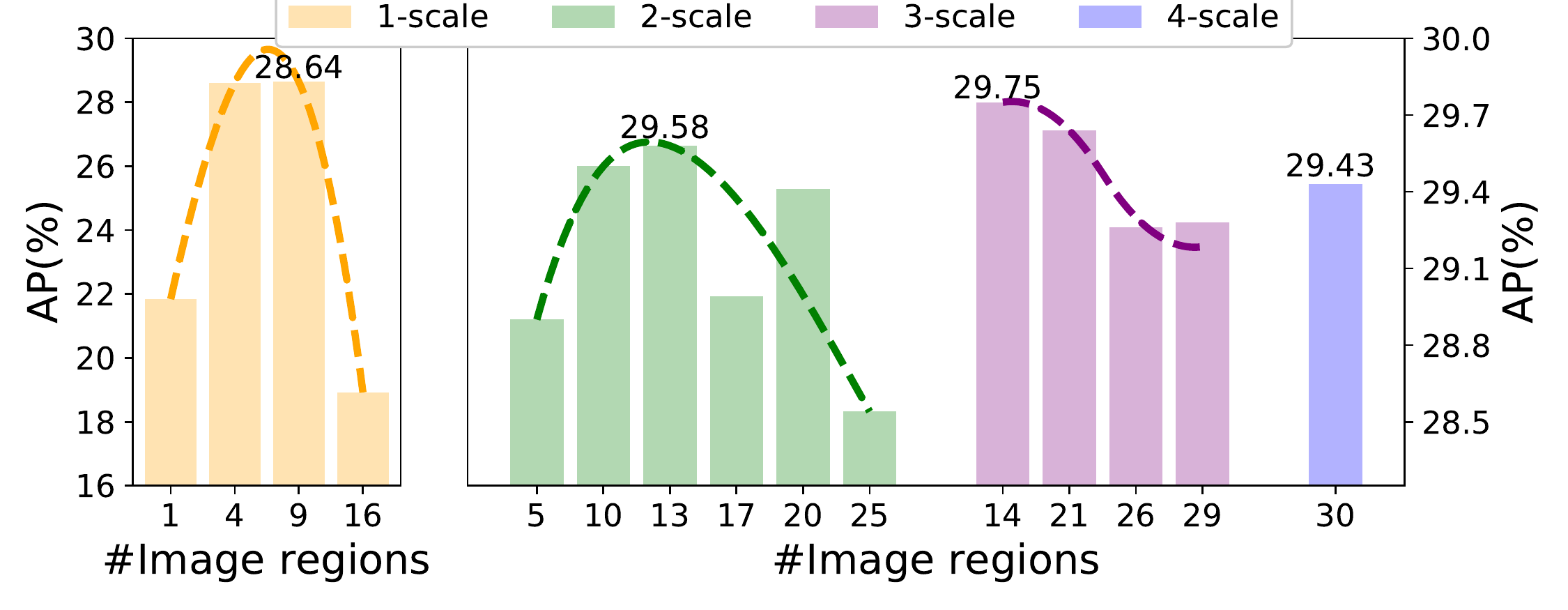}

  \caption{The average precision of uniform partition with different scales and their combinations on VisDrone2019 dataset. The UP of
  appropriate settings improves the AP and multi-scale UP further improves the detection accuracy. However, when the number of regions
   increases, the detection accuracy drops notably. }

  \label{UPmAP}
\end{figure}

For effective object detection in large-scale scenes, it is essential to focus on the object-centric regions. The Uniform Partition (UP) is a straightforward strategy to ``focus" on image regions in a sliding-window way. It partitions an image into several uniform regions and enlarges those regions for detection. We first conduct the UP strategy on the VisDrone2019~\cite{zhu2018vision} for analysis. Various settings of UP $(1\times 1,\ 2\times 2,\ 3\times 3,\ 4\times 4)$ and all the combinations are evaluated ($n\times m$ denotes uniformly partitioning the image into $n\times m$ regions). The detection accuracy measured by average precision (AP) is presented in Fig.~\ref{UPmAP}. The UP of appropriate settings does improve the AP, which validates the importance of focus regions in large scenes. Compared to single-scale setting (Fig.~\ref{UPmAP} left), multi-scale UP would further improve the detection accuracy (Fig.~\ref{UPmAP} right). However, when the number of regions increases, the detection accuracy drops notably. That is because the false negatives accumulate with the multi-scale repeated partitioned regions. Besides, with more cropped regions, the possibility of objects truncated by additional cropping is higher, which would cause more repeated and incomplete detections. Moreover, the UP strategy enlarges the cropped image regions with a fixed scale, which is inappropriate for scale-adaptive object detection in large scenes. In a word, the performance of UP strategy is limited by the lacking of adaptivity issue.

\begin{figure*}[ht]
  \centering
  \includegraphics[width=1.0\textwidth]{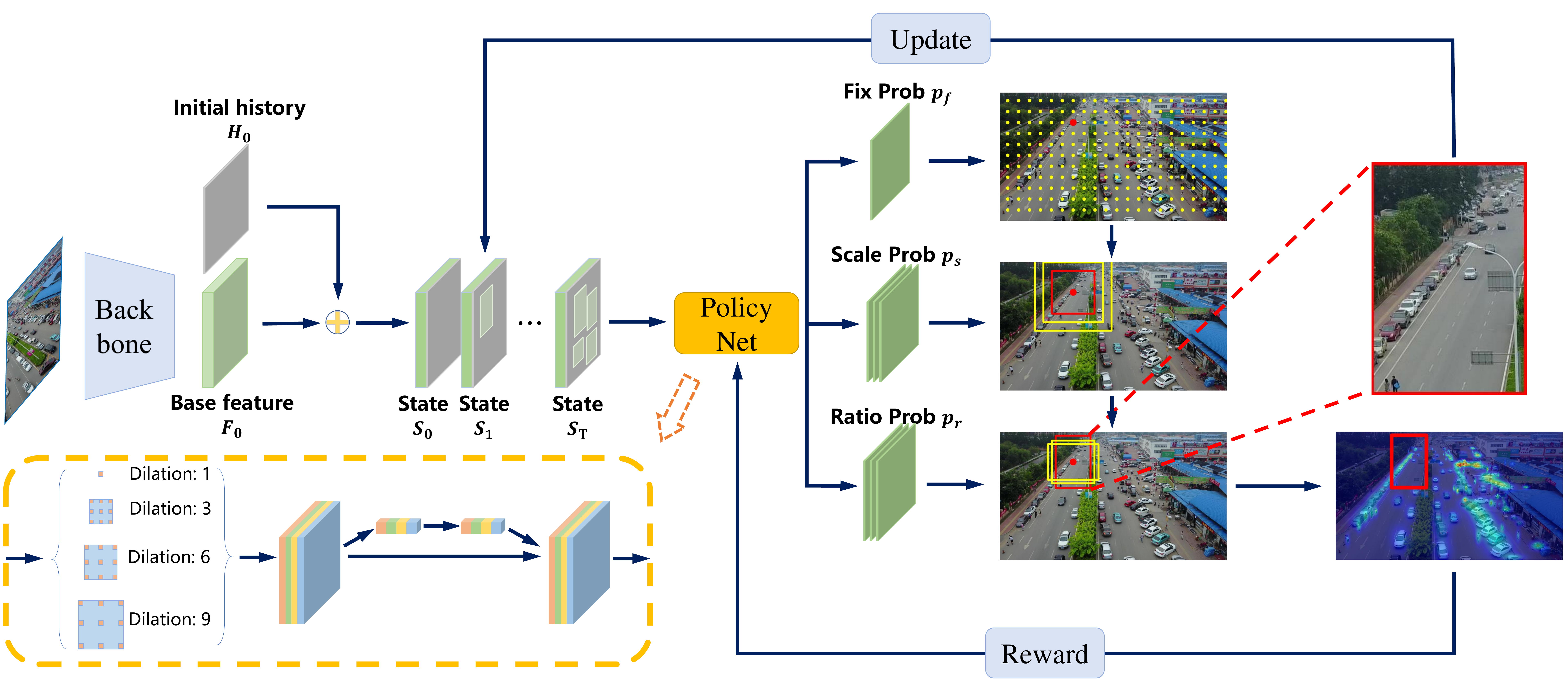}
  \caption{The workflow of AdaZoom. The initial state $S_0$ is composed of base feature $F_0$ and history information map $H_0$. $F_0$ is extracted from the image and the values of $H_0$ are all set to zero. The PolicyNet generates three branches for the probability of fixation, scale, and aspect ratio, respectively. The focus region is sampled from these probability maps. Then the state $S_1$ is updated for the next time step according to the focus region. The reward of the focus region is derived from the object distribution.}\label{AdaZoom}
\end{figure*}

To tackle this, it is essential to adaptively generate focus regions in large-scene images for effective and efficient detection. Yet it is difficult to label the focus regions as true or false since the semantics is ambiguous. This makes it difficult to learn the designed network in supervised learning scheme. Based on these considerations, we propose an Adaptive Zoom network (AdaZoom) based on reinforcement learning (RL), as shown in Fig.~\ref{AdaZoom}. We design a continuous reward to measure the quality of focus regions based on the scales and distribution of objects. The RL agent is then encouraged to explore an adaptive region generation policy which maximizes the accumulated reward. In particular, AdaZoom can localize and zoom the focus regions with flexible shape and scale. The focus regions are adaptive to the scales and distribution of objects, which can significantly alleviate the scale variation towards high-performance object detection in large scenes.

\subsection{Problem Formulation}

\begin{figure}[h]
  \centering
  \includegraphics[width=1.0\linewidth]{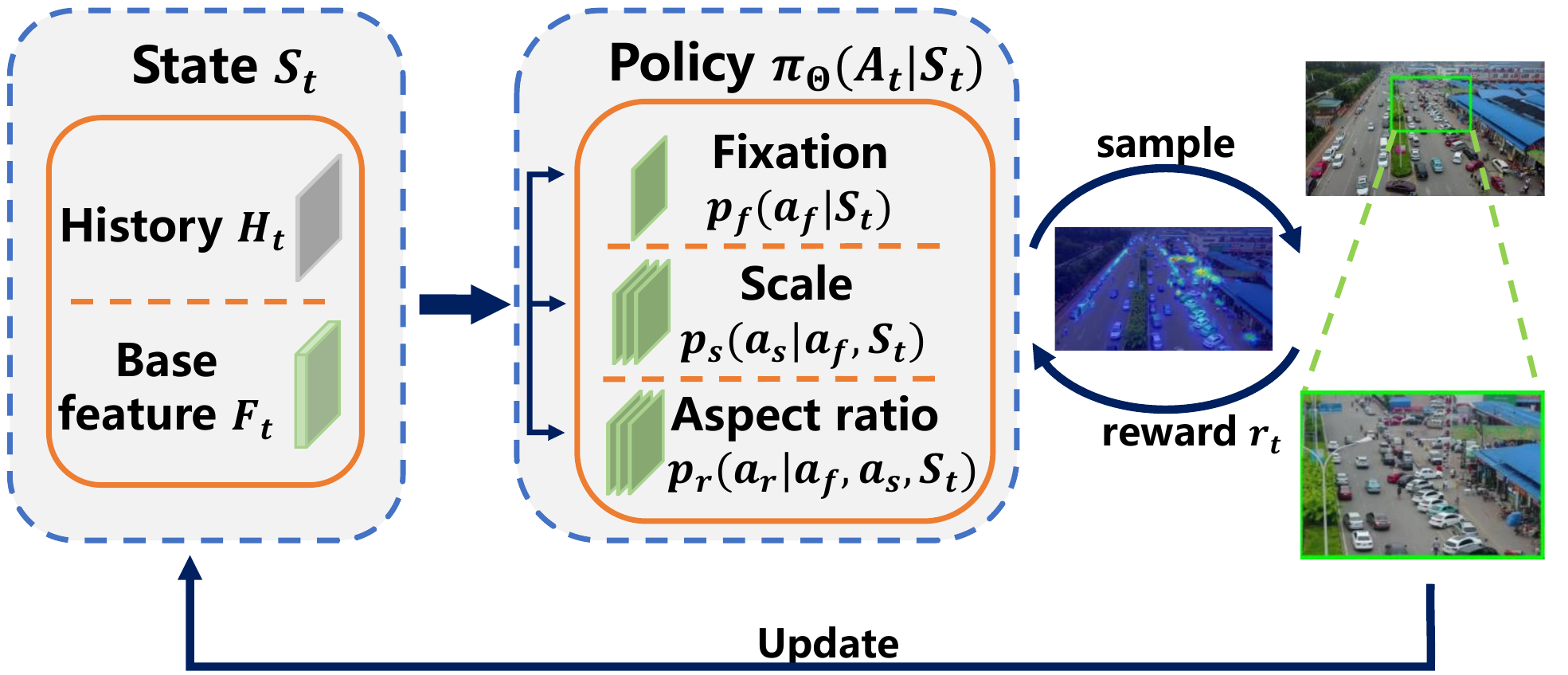}
  \caption{Formulation of focus region generation. The state $S_t$ is composed of base feature $F_t$ and history information map $H_t$. The action $A_t$ is decoupled to the fixation $a_f$, scale $a_s$ and aspect ratio $a_r$ of focus regions. Object distribution is introduced into the sampling process. The reward is derived from object distribution.}
  \label{formulation}
\end{figure}

We construct AdaZoom with reinforcement learning framework. Based on policy gradient~\cite{1999Policy}, AdaZoom is optimized according to the reward which measures the quality of the focus region. We view the sequence of region generation as a Markov Decision Process (MDP)~\cite{bellman1957markovian} and formulate adaptive focus region generation as a reinforcement learning problem. As shown in Fig.~\ref{formulation},  we construct the Policy Network $\pi_{\Theta}$ to generate the probability distribution of action space based on the state $S_t$. The action $A_t$ is decoupled into fixation, scale, and aspect ratio of the focus region. The object distribution is referred to guide the sampling process for better convergence. Then we derive the reward $r_t$ from object distribution to measure the quality of a region. The state $S_{t+1}$ is updated according to the generated region. For each image, $T$ time steps make up an episode and we optimize the Adazoom with policy gradient~\cite{1999Policy} to maximize the expected cumulative reward of each episode.
%We set $\gamma$ to 0.5. The gradient of the expected cumulative reward is calculated by:
%\begin{equation}\label{eq3}
%\begin{split}
%   \nabla_{\Theta}J(\Theta_k) = \mathbb{E}_{\pi_\Theta}\left[\sum_{t=0}^{T}\gamma^t\nabla_{\Theta}\log\pi_\Theta(A_t|S_t)r_t\right]
%\end{split}
%\end{equation}
%------------------------------------------------------------------------
%\subsubsection{State}

\vspace{10.0pt}
\textbf{State.} The state $S_t$ consists of the base feature map $F_t$ and the history information map $H_t$. $F_t$ is supposed to learn the object-wise scale and distribution information at a coarse level. It is distinct from the feature maps for object detection which focus on the fine-level features of each object. The binary history information map $H_t$ records the regions generated at previous steps. The initial state $S_0$ is the concatenation of an all-zero binary map $H_0$ and the base feature map $F_0$ which is extracted from the image by a backbone network. The focus region is mapped from the image to the state, denoted as $z_t$. The state is updated as follows:
%\begin{equation}\label{ep5}
%    F_{t+1}(i,j) = \left\{
%              \begin{array}{lc}
%                \kappa F_{t}(i,j), & if \ (i,j)\in z_t \\
%                \\
%                F_{t}(i,j), & otherwise
%              \end{array}
%              \right.
%\end{equation}
\begin{equation}\label{ep4}
\footnotesize
  H_{t+1}(i,j) = \mathbb{I}\{(i,j)\in z_t\} + H_{t}(i,j)\mathbb{I}\{(i,j)\notin z_t\}
\end{equation}
\begin{equation}\label{ep5}
\footnotesize
  F_{t+1}(i,j) = \kappa F_{t}(i,j)\mathbb{I}\{(i,j)\in z_t\} + F_{t}(i,j)\mathbb{I}\{(i,j)\notin z_t\}
\end{equation}

% \begin{equation}\label{ep4}
%    H_{t+1}(i,j) = \left\{
%              \begin{array}{lc}
%                1, & if \ (i,j)\in z_t \\
%                \\
%                max(0,\ H_{t}(i,j)), & otherwise
%              \end{array}
%              \right.
%\end{equation}
\noindent where $\mathbb{I}(\cdot)$ is the indicator function and $\kappa$ is a decay factor to suppress the response in the corresponding focus region. We set $\kappa$ to 0.1.

%\subsubsection{Action}
\vspace{10.0pt}
\textbf{Action.} The action $A_t$ is sampled from the probability distribution $\pi_\Theta(A_t|S_t)$. We design a three-branch Policy Network to progressively learn the fixation, scale and aspect ratio of focus regions. The first branch generates the probability distribution for fixation $p_f(a_f|S_t)$, where $a_f$ is a point on the fixation probability map $p_f\in\mathbb{R}^{h\times w}$. Inspired by the anchor based mechanism~\cite{ren2015faster}, each point in the fixation probability map represents a region center location in the image. The second branch generates a scale probability map $p_s\in\mathbb{R}^{h\times w \times{n_s}}$, where $n_s$ is the predefined number of candidate region scales. $p_s(a_s|a_f; S_t)$ is conditioned on the fixation $a_f$, where $a_s$ is the region scale. The third branch generates aspect ratio probability map $p_r\in\mathbb{R}^{h\times w\times{n_s}\times{n_r}}$, where $n_r$ is the predefined number of candidate aspect ratios. The aspect ratio probability distribution $p_r(a_r|a_f,a_s;S_t)$ is conditioned on the fixation $a_f$ and scale $a_s$. The action $A_t$ is composed of $(a_f, a_s, a_r)$ to specify a region. We formulate the policy as follows:
\begin{equation}\label{ep6}
\footnotesize
    \pi_\Theta(A_t|S_t)=p_f(a_f|S_t)\times{p_s(a_s|a_f;S_t)}\times{p_r(a_r|a_f,a_s;S_t)}
\end{equation}

These branches coordinate with each other by focusing on region representation from different points of view. The fixation branch tries to find the center of a cluster of objects. The scale branch is supposed to adjust the scale of the region according to the scales of objects around the fixation. The aspect ratio branch adapts to the distribution of objects around the fixation with a selected scale of the region. The network structure is detailed in the Supplementary Material.

%We adopt CSRNet~\cite{li2018csrnet}, a network designed for dense scenes, as the backbone to extract the feature for describing the object scale and distribution as the state. The Policy Network is shown in Fig.~\ref{AdaZoom}. We adopt ASPP~\cite{chen2017deeplab} and SENet~\cite{hu2018squeeze} to combine feature maps with different receptive field. The fixation, scale and aspect ratio branches adopt deformable convnets~\cite{zhu2019deformable} following SENet~\cite{hu2018squeeze}. As a consequence, the Policy Network is powerful in extracting features of object-wise scale and distribution information for focus region generation.

%\subsection{Reward}
%\subsubsection{Reward}
\vspace{10.0pt}
\textbf{Reward.} There are no clear annotations to supervise the region generation since the image can be partitioned into reasonable regions in many ways and the semantics of the region is ambiguous for supervised learning. Therefore, we design the reward which measures the quality of the regions. The reward is derived from annotations for bounding boxes of objects. AdaZoom is expected to pay more attention to small objects. For the $i_{th}$ object, it is assigned a weight $w_i \propto \frac{1}{s_i}$, where $s_i$ is the scale of the object. For each scale $a_s$ of region, there is a desired object scale range $[a_s^{min}, a_s^{max}]$. The reward at step $t$ is defined as follows:
\begin{equation}\label{ep7}
    r_t(A_t) = \frac{\sum_{Z_t}{I_i(a_s) w_i}}{\sum_{\hat{Z}}w_j}
\end{equation}
where $Z_t$ is the set of objects enclosed in the $t_{th}$ focus region and $\hat{Z}$ corresponds to the remaining regions of the image until step $t-1$. In general, $r_t$ can be regarded as a weighted recall of object weights $w$ at step $t$. $I_i(a_s)$ denotes the measurement of consistency between scales of objects and regions. For the $i_{th}$ object in the focus region with $a_s$ scale, if the scale of the object falls beyond $[a_s^{min}, a_s^{max}]$, $w_i$ will decay by $I_i(a_s)$ and the mismatch between the scale of the region and object is introduced to reward. $I_i(a_s)$ is defined as follows:
\begin{equation}\label{ep8}
\footnotesize
    I_i(a_s) = \left\{
              \begin{array}{cl}
                1, & \ s_i \in [a_s^{min},\  a_s^{max}] \\
                \\
                max(0,\ 2-e^{\beta\Delta s}), & otherwise
              \end{array}
              \right.
%\footnotesize
%    I_i(a_s) = \mathbb{I}\{(i,j)\in [a_s^{min},\  a_s^{max}]\} + \mathbb{I}\{(i,j)\notin [a_s^{min},\  a_s^{max}]\}max(0,\ 2-e^{\beta\Delta s})
\end{equation}
\noindent where $s_i$ is the scale of the object, $\Delta s= \frac{|s_i-a_s^{min(max)}|}{a_s^{min(max)}}$ is the fraction that $s_i$ exceeds $a_s^{max}$ or smaller than $a_s^{min}$. $\beta$ is a positive coefficient to adjust the decay rate for scales beyond scale range. we set $\beta$ to 1.5.
%-------------------------------------------------------------------------
\subsection{Training and Inference}
Our approach for object detection in large-scale scenes contains two components: (1) AdaZoom network to adaptively zoom the focus regions; (2) Detection network to locate the objects in focused regions. We train AdaZoom and detection network collaboratively to boost the performance.

%\subsection{Sampling and Training}
%\textbf{Sampling.} The focus region generated by AdaZoom is parameterized by fixation, scales, and aspect ratio. The dimension of the action space is $h\times w\times{n_s}\times{n_r}$. Such large action space is difficult to explore even the Policy Network fails to converge. To tackle this, we decouple the action $A_t$ into three sub-actions: fixation $a_f$, scale $a_s$ and aspect ratio $a_r$, based on the formulation of policy Eq.~\ref{ep6}. We firstly sample a fixation $a_f$ from the fixation probability map $p_f$.
%With the sampled fixation $a_f$, we greedily search from the candidate scales and aspect ratios to maximize the reward at fixation $a_f$ to produce the pseudo labels for $a_s$ and $a_r$. Then, scale and aspect ratio branches are optimized to minimize the focal loss~\cite{lin2017focal} between the probability of $a_s$, $a_r$ and the pseudo labels. As the consequence, the dimension of action space reduces to $h\times w$, which helps to converge to a satisfying optimal policy.

\begin{figure}[t]
  \centering
  \includegraphics[width=.47\textwidth]{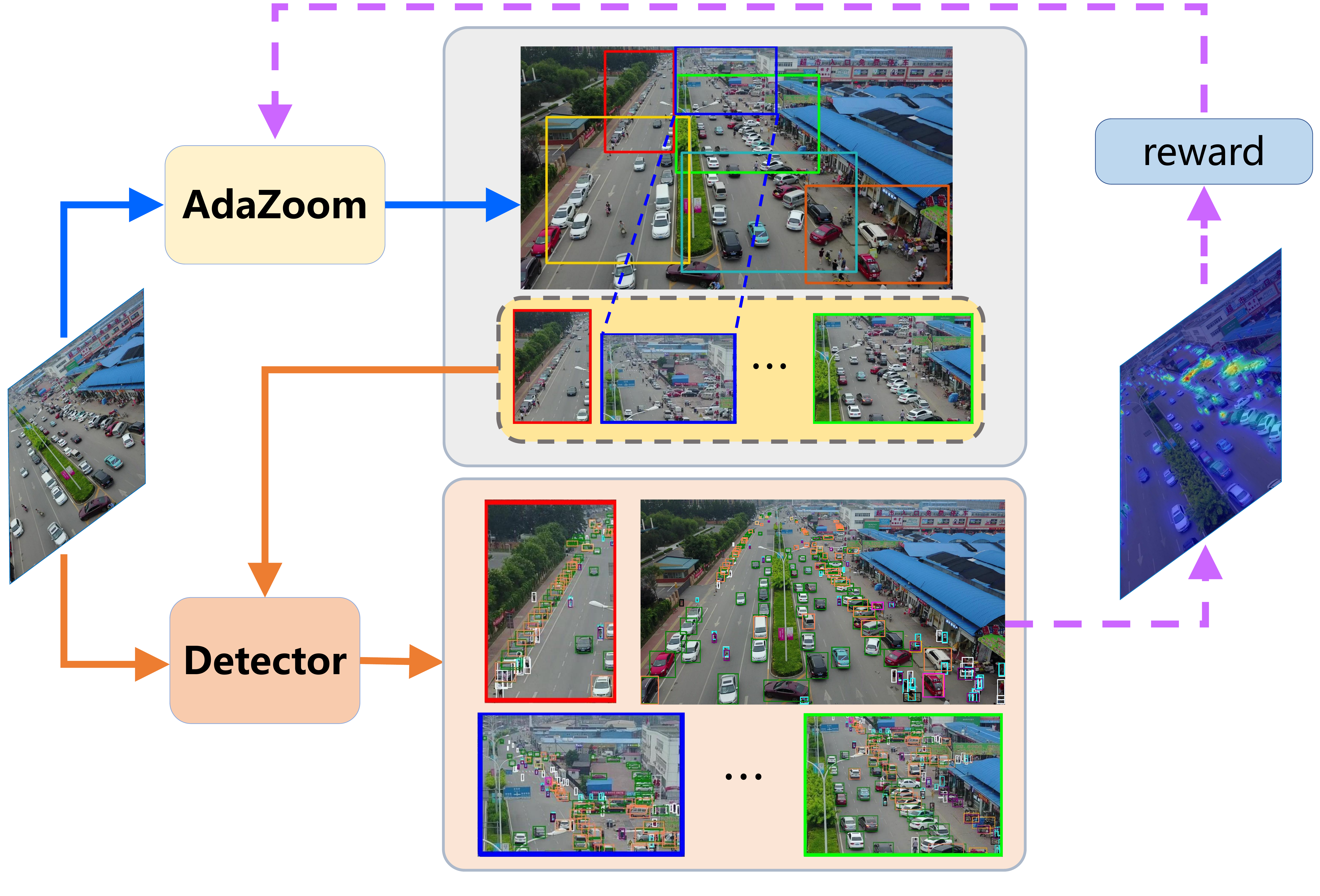}
  \caption{When training the detector, the focus regions generated by AdaZoom are resized to a certain scale to finetune the detector. When training AdaZoom, the detection results of the image are introduced to the reward for AdaZoom. During inference, the final detection results are merged from focus regions and the image.}\label{joingly}
\end{figure}

\textbf{Collaborative training.} AdaZoom generates a series of focus regions with different scales and aspect ratios. We crop these focus regions from the original image and resize them to a certain scale so that smaller regions obtain higher magnification. In order to alleviate the domain shift between original images and resized regions, the detector is re-trained on the resized regions. Different from the existing work~\cite{gao2018dynamic, najibi2019autofocus,singh2018sniper}, we integrate the training and inference in the same pipeline with the focus regions generated by AdaZoom. In addition, we design collaborative training to further improve the performance of detection. During collaborative training, AdaZoom is expected to focus on the regions difficult for the re-trained detector. The detector infers on the image and outputs confidence scores $c_i$ for each object. For false negative, the confidence score is set to zero. Then the weight $w_i$ in Eq.~\ref{ep7} is modified as $w_i=1-c_i$. The AdaZoom is trained on the new modified weights. As a consequence, AdaZoom pay attention to difficult regions where the confidence of true positive is low. Then the difficult regions generated by the AdaZoom are used to finetune the detector (Fig.~\ref{joingly}). This simple modification promotes the coordination between AdaZoom and the detector.

\textbf{Inference.} Inference shares the same pipeline with training. We adopt the greedy sampling to take actions:
\begin{equation}\label{ep10}
\left\{
  \begin{aligned}
    a_f = & \mathop{\arg\max}_{a_f}p_f(a_f|S_t) \\
    a_s =& \mathop{\arg\max}_{a_s}p_s(a_s|a_f; S_t) \\
    a_r =& \mathop{\arg\max}_{a_r}p_r(a_r|a_f, a_s; S_t)
  \end{aligned}
\right.
\end{equation}

The fixation $a_f$ is selected as the center of regions. Conditioned on fixation $a_f$, AdaZoom selects a scale $a_s$ that implies the scales of objects around fixation. With the selected fixation $a_f$ and scale $a_s$, an aspect ratio $a_r$ is selected to adapt to the distribution of objects around fixation. Based on the selected actions, a region is generated. The generated regions as well as the original image are resized together to a certain scale as a batch for detection. The final results of each region are merged together by non-maximum suppression(NMS) with the IoU threshold setting to 0.5.

%\vspace{-0.2cm}
%-------------------------------------------------------------------------
\section{Experiments}
\subsection{Dataset and Metric}
We conduct experiments on the public detection benchmark VisDrone2019~\cite{zhu2018vision}, UAVDT~\cite{du2018unmanned} and DOTA~\cite{xia2018dota} to evaluate our method. (1) \textbf{VisDrone2019}~\cite{zhu2018vision} consists of 10,209 images for detection task with \textit{train} set of 6,471 images, \textit{val} set of 548 images. The \textit{test} set is split into \textit{test-dev} with 1,610 images and \textit{test-challenge} with 1,580 images. The resolution of images can be as large as $2000\times1500$ pixels. On average, each image has 53 objects and most of them are small as well as densely distributed. (2) \textbf{UAVDT}~\cite{du2018unmanned} is another popular UAV-based detection benchmark. It consists of 23258 training images and 15069 testing images. The average resolution of images is about $1080\times540$ pixels. The tiny objects just contain 0.005\% pixels of a frame. (3) \textbf{DOTA}~\cite{xia2018dota} is a public dataset for remote sensing. There are 1411 images for training and 458 images for validation. Following ClusDet~\cite{yang2019clustered},  we choose the images with movable objects such as plane, ship, vehicle and helicopter. The dataset contains 920 images for training and 285 images for validation.

We follow the evaluation COCO-style protocol in~\cite{zhu2018vision} with the official evaluation toolkit \footnote{https://github.com/VisDrone/VisDrone2018-DET-toolkit} for VisDrone2019. We also use the evaluation protocol in COCO~\cite{lin2014microsoft} for UAVDT and DOTA. The detection performance is evaluated with metrics of $AP^{[0.5:0.05:0.95]}$, $AP^{0.5}$, $AP^{0.75}$.

\begin{table}[h]\footnotesize
\begin{center}
\setlength{\tabcolsep}{3.0pt}
\begin{tabular}{c|c|ccc|c}
  \hline
  Dataset                    &Method       & $AP$ & $AP_{50}$ & $AP_{75}$ & s/img(GPU)  \\
\hline\hline
  %\multirow{6}*{\tabincell{c}{VisDrone2019\\ \textit{test-dev}}}
  \multirow{6}*{VisDrone2019}
                            &UP($1\times1$)      & 21.84  & 40.92  & 21.46  & 0.070 \\
                            &UP($2\times2$)      & 28.61   & 51.97 & 28.34 & 0.270 \\
                            &UP($3\times3$)      & 28.64   & 52.70  &27.90  & 0.617 \\
                            \cline{2-6}
                            &Multi-Ratio UP      & 29.01   & 52.99   & 28.46  & 1.078\\
                            &Multi-Scale UP      & 29.75   & 53.89  & 29.56 & 0.960\\
                            \cline{2-6}
                            &AdaZoom             & \textbf{31.22}   & \textbf{56.16}   & \textbf{31.22} & 0.654\\

\hline
  \multirow{6}*{UAVDT}       &UP($1\times1$)      & 12.1  &23.5   &10.8  & 0.067 \\
                             &UP($2\times2$)    &13.3   & 25.1  & 13.1  & 0.252\\
                             &UP($3\times3$)    &10.9   & 21.3  & 9.9  & 0.558\\
                             \cline{2-6}
                             &Multi-Ratio UP      & 15.0   & 27.3   & 15.3 & 1.008\\
                             &Multi-Scale UP      & 15.3   & 28.1  & 15.4 & 0.854\\
                             \cline{2-6}
                             &AdaZoom             & \textbf{19.6}   & \textbf{33.6}   & \textbf{21.3} &0.599\\

\hline
\end{tabular}
\end{center}
\caption{Comparison with uniform partition on VisDrone2019 \textit{test-dev} and UAVDT \textit{test} dataset. We adopt Faster R-CNN as the detector. We also report the inference time per image.}
\label{up}
\end{table}

\begin{table}
\begin{center}
\setlength{\tabcolsep}{8.5pt}
\footnotesize
\begin{tabular}{c|c|ccc}
  \hline
   Method     &Backbone  & $AP$ & $AP_{50}$ & $AP_{75}$  \\
\hline\hline

 Faster R-CNN~\cite{lin2017feature}  &  ResNet50   & 12.1   & 23.5  & 10.8 \\
 ClusDet~\cite{yang2019clustered}  &   ResNet50    & 13.7   & 26.5  & 12.5  \\
 DMNet~\cite{li2020density}        & ResNet50    &14.7 &24.6 &16.3  \\
 DREN~\cite{zhang2019fully}        & ResNet50    &15.1 &-&- \\
 GLSAN~\cite{deng2020global}       & ResNet50     & 17.0 & 28.1 & 18.8 \\
 GLSAN$\dagger$~\cite{deng2020global}       & ResNet50     & 19.0 & 30.5 & 21.7  \\
 AdaZoom               &   ResNet50    &19.6    &33.6    &21.3    \\
 AdaZoom$\dagger$               &   ResNet50    &\textbf{22.4}    &\textbf{38.6}    &\textbf{23.9}   \\
 \hline
 Faster R-CNN~\cite{lin2017feature}  &  ResNet101   & 15.1   & 26.5  & 16.0  \\
 GLSAN~\cite{deng2020global}       & ResNet101     & 17.1 & 28.3 & 18.8 \\
 DREN~\cite{zhang2019fully}        & ResNet101    &17.7 &-&- \\
 AdaZoom             &   ResNet101    &\textbf{20.1}    &\textbf{34.5}    &\textbf{21.5}    \\
\hline
\end{tabular}
\end{center}
\caption{Detection performance on UAVDT \textit{test} dataset. We adopt Faster R-CNN and Cascade R-CNN as the detector. † denotes Cascade R-CNN.}
\label{UAVDT}
\end{table}

\subsection{Implementation Details}
\label{training detail}
For AdaZoom, we set $n_s=3$ for scale candidates $[240^2,350^2,420^2]$ and the desired scale ranges are set to $(0,40^2)$,$(30^2,60^2)$,$(50^2, \infty)$, respectively. The candidate aspect ratios are $[0,7, 1.0, 1.5]$. %The $\alpha$ in focal loss is set to (0.1, 0.3, 0.6) for scale branch and (0.3,0.4,0.3) for aspect ratio branch.
The max step number per episode $T$ is set to 7, empirically. We train the AdaZoom with 16 batch size and 2e-5 learning rate for 5k iterations.
Fot the detector (\ie, Faster R-CNN~\cite{ren2015faster} and Cascade R-CNN~\cite{cai2018cascade}), we follow the default configurations of maskrcnn-benchmark~\cite{massa2018mrcnn}. We train the detector for 90k iterations with 0.001 learning rate. The learning rate is decreasing by a factor of 0.1 after 60k iterations and 80k iterations. During training and inference, the short edge of regions generated by AdaZoom are resized to 800 pixels.
For collaborative training, we iteratively finetune AdaZoom for 500 iterations and detector for 1000 iterations with learning rate decayed by 0.1.

\subsection{Comparison with Baseline}

\begin{table}[ht]
\begin{center}
\setlength{\tabcolsep}{4.pt}
\footnotesize
\begin{tabular}{l|c|ccc}
  \hline
  Method         & Backbone & $AP[\%]$ & $AP_{50}[\%]$  & $AP_{75}[\%]$    \\
\hline\hline
  RRNET~\cite{chen2019rrnet} &Hourglass&32.92 & -  & 31.33 \\
  CRENet~\cite{wang2020object}  & Hourglass-104 &33.70  & 54.30  & 33.50  \\
  DMNet~\cite{li2020density} & ResNet50   & 28.20  &  47.60 & 28.90  \\
  CascadeNet~\cite{zhang2019dense} &ResNet50  &30.12  &58.02 & 27.53 \\
  GLSAN~\cite{deng2020global}     &ResNet50   &30.70  &55.40  & 30.00   \\
  SAMFR~\cite{wang2019spatial} &ResNet50  &33.72  &58.62 & 33.88 \\
  MPFPN~\cite{liu2020small}  &ResNet101 & 29.05 & 54.38 & 26.99  \\
  GLSAN~\cite{deng2020global} & ResNet101 &30.70  & 55.60  & 29.90 \\
  ClusDet~\cite{yang2019clustered} & ResNeXt101 &32.40  &56.20  &31.60 \\
  SAIC-FPN~\cite{zhou2019scale} &ResNeXt101 & 35.69 &62.97 &35.08 \\
  DREN~\cite{zhang2019fully} & ResNeXt152 &30.30 &-&- \\
\hline
  w/o CT:  & &&& \\
  AdaZoom                  & ResNet50    & 34.71  & 62.16  & 33.89    \\
  AdaZoom                  & ResNeXt101 & 36.56  & 64.58  & 36.03   \\
  AdaZoom$\dagger$                  & ResNeXt101 & 38.69  & 65.35  & 39.45  \\
\hline
  w/ CT:  & &&& \\
  AdaZoom                  & ResNet50    & 36.19  & 63.50  & 36.11 \\
  AdaZoom                 & ResNeXt101 & 37.58  & 66.25  & 37.34  \\
  AdaZoom$\dagger$               & ResNeXt101 & \textbf{40.33}  & \textbf{66.94}  & \textbf{41.77}  \\
\hline
\end{tabular}
\end{center}
\caption{Detection performance on VisDrone2019 \textit{val} dataset. Results for SOTA are taken from the publications. We adopt Faster R-CNN and Cascade R-CNN as the detector following AdaZoom and $\dagger$ denotes Cascade R-CNN. We report the results with and without collaborative training (CT), respectively.}
\label{VisdroneSOTA}
\end{table}

We first compare the proposed AdaZoom with the Uniform Partitions (UP) as the baseline, to evaluate the effectiveness. For UP, a whole image is uniformly partitioned into $m\times n$ regions with 50 pixels overlap. In particular, we implement comprehensive experiments of UP to provide a simple yet strong baseline. We evaluate multi-scale UP of $[1\times 1,\ 2\times2,\ 3\times3]$ and multi-ratio UP of $[2\times 3,\ 2\times2,\ 3\times2]$. Table.~\ref{up} shows the comparative study with Faster R-CNN as the detector and ResNet-50 backbone. On VisDrone2019 dataset, the AdaZoom achieves $AP$ of 31.22\% and $AP_{50}$ of 56.16\%. Compared to UP of $3\times 3$, the AdaZoom improves $AP$ by $2.58\%$ with comparable inference time. Compared to the multi-scale UP and multi-ratio UP, the AdaZoom improves the $AP$ by $1.47\% \sim 2.21\%$ and $AP_{50}$ by $2.27\% \sim 3.17\%$. On UAVDT dataset, the AdaZoom achieves $AP$ of 19.6\% and $AP_{50}$ of 33.6\%, outperforming UP by a large margin. Compared to multi-scale UP and multi-ratio UP, we improve $AP$ by $4.3\% \sim 4.6\%$ and $AP_{50}$ by $5.5\% \sim 6.3\%$, respectively. Meanwhile, compared to multi-scale UP, we save the $31.8\%$ and $29.8\%$ of inference time on VisDrone2019 and UAVDT dataset, respectively. The results validate that the proposed AdaZoom is both effective and efficient of object detection in large scenes. That is mainly because our method generates focus regions in an adaptive way. Instead of sliding over the whole image, we only generate the important regions with appropriate scales and shapes. Therefore, our method is adaptive to scale varying. Besides, the AdaZoom would avoid generating too many regions, further leading to notable accuracy boosting and inference time saving.

\subsection{Comparison with State-of-The-Art Models}

We compare our method with the state-of-the-art methods across a wide range of datasets, such as UAVDT \textit{test} (Table.~\ref{UAVDT}), VisDrone2019 \textit{val} (Table.~\ref{VisdroneSOTA}) and DOTA \textit{val} dataset (Table.~\ref{DOTA}). From the tables we can see that our method achieves superior performance over existing methods on all the three datasets. On UAVDT dataset, we achieve $22.4\%$ of $AP$ and $38.6\%$ of $AP_{50}$ with Cascade R-CNN as the detector. Compared to GLSAN~\cite{deng2020global} with Cascade R-CNN, the $AP$ is increased by 3.4\% and $AP_{50}$ is increased by 8.1\%. On Visdrone dataset, the AdaZoom achieves 40.33\% and $AP_{50}$ of 66.94\%, outperforming the SOTA performance by a large margin. On DOTA dataset, we follow the experimental setting as~\cite{yang2019clustered} to evaluate the detection performance in large scenes. We achieve 37.8\% of $AP$ and 63.5\% of $AP_{50}$, with ResNeXt-101 as backbone and Faster R-CNN as the detector. The achieved $AP$ is increased by 5.5\% $\sim$ 6.4\%. We also focus on the comparison with the existing methods which apply the cropped and resized regions for detection boosting, as follows.

\textbf{Comparison with pseudo annotation based region generation methods.} For such methods, pseudo annotations of focus regions are produced with the object annotations. Typically, ClusDet~\cite{yang2019clustered} trains a region generation network by supervised learning based on pseudo generated annotations. For fair comparison, we use the Faster R-CNN with the same backbone with the ClusDet. On Visdrone dataset, our method outperforms ClusDet by $4.16\%$ in $AP$ even without collaborative training. It is reasonable since the region selection is difficult to be formulated as a supervised learning problem. Assigning the hard labels as true or false to focus regions is semantically vague. In contrast, we design a continuous reward to measure whether a generated region is good or bad. The RL formulation is more suitable for such problem. In addition, by integrating collaborative training, our method outperforms ClusDet in $AP$ by $5.18\%$ on VisDrone2019 and $5.9\%$ on UAVDT dataset. On DOTA~\cite{xia2018dota} dataset, we set the number of focus regions of AdaZoom to 3. It should be noted from Table.~\ref{DOTA} that our method achieves consistent improvement with stronger backbone.

\begin{table}[t]
\begin{center}
\setlength{\tabcolsep}{7.9pt}
\footnotesize
\begin{tabular}{c|c|ccc}
  \hline
   Method     &Backbone  & $AP$ & $AP_{50}$ & $AP_{75}$  \\
\hline\hline

 Faster R-CNN~\cite{lin2017feature}  &  ResNet50   & 30.60   & 52.10  & 31.30 \\
 ClusDet~\cite{yang2019clustered}  &   ResNet50    & 32.20   & 47.60  & \textbf{39.20}  \\
 AdaZoom               &   ResNet50    &\textbf{36.00}    &\textbf{62.70}    &37.00    \\
 \hline
 Faster R-CNN~\cite{lin2017feature}  & ResNet101   & 30.50   & 52.10  & 31.00   \\
 ClusDet~\cite{yang2019clustered}  &   ResNet101    & 31.60   & 47.80  & \textbf{38.20}  \\
 AdaZoom              &   ResNet101    &\textbf{36.10}    &\textbf{63.10}    &36.20    \\
 \hline
 Faster R-CNN~\cite{lin2017feature}  & ResNeXt101   & 32.30   & 54.50  & 33.30 \\
 ClusDet~\cite{yang2019clustered}  &   ResNeXt101    & 31.40   & 47.10  & 37.40 \\
 AdaZoom               &   ResNeXt101    &\textbf{37.80}    &\textbf{63.50}    &\textbf{39.20}    \\

\hline
\end{tabular}
\end{center}
\caption{Detection performance on DOTA \textit{val} dataset. We adopt Faster R-CNN as detector and employ different backbones.}
\label{DOTA}
\end{table}

\textbf{Comparison with coarse-to-fine detection methods.} For such methods, the object distribution is estimated based on a coarse-level preview detection, such as DREN~\cite{zhang2019fully}, CRENet~\cite{wang2020object} and GLSAN~\cite{deng2020global}. Compared to CRENet with backbone of Hourglass-104, Our AdaZoom with Faster RCNN achieves $2.49\%$ higher in $AP$ on VisDrone2019 with  ResNet-50 backbone. Compared to DREN~\cite{zhang2019fully} and ResNet101, the $AP$ is increased by $4.5\%$ with ResNet-50 backbone and $2.4\%$ with ResNet-101 backbone on UAVDT dataset. Compared with GLSAN~\cite{deng2020global}, which uses extra super-resolution network for enlarging regions, our method outperforms it by $5.49\%$ of $AP$ on VisDrone2019. The performance of the coarse-to-fine methods is limited by the initial detection, which would cause the bias for region generation (\textit{i.e.}, small objects are easily missed in coarse-level detection). In contrast, our AdaZoom cooperates well with detector. The performance of both AdaZoom and detector are improved by collaborative training.

\begin{table}[t]
\begin{center}
\footnotesize
\setlength{\tabcolsep}{9.9pt}
\begin{tabular}{cc|ccc}
  \hline
    SR &CT& $AP[\%]$ & $AP_{50}[\%]$  & $AP_{75}[\%]$    \\
\hline\hline

                                &    & 29.74 & 54.06  & 29.48 \\

                        & \checkmark & 29.99 & 54.21  & 29.86   \\

             \checkmark& & 30.51  & 55.28   & 30.22  \\

         \checkmark&\checkmark& \textbf{31.22}  & \textbf{56.16}  & \textbf{31.22}  \\

\hline
\end{tabular}
\end{center}
\caption{Ablation for AdaZoom on VisDrone2019 \textit{test-dev}. We adopt Faster R-CNN with ResNet50 as detector. \textbf{SR}: Scale and ratio adaptation. \textbf{CT}: collaborative training.}
\label{ablation}
\end{table}

%\textbf{Region Generation Based on Clustering.} DMNet~\cite{li2020density} design the network to predict the density map for the image. The focus regions are generated by clustering on the density map.

\begin{figure}[t]
%\includegraphics[width=.5\textwidth]{pic/recall.pdf}
%    \subcaptionbox{\includegraphics[width=.5\linewidth]{pic/recall.pdf}}\hfill
%    \subcaptionbox{\includegraphics[width=.5\linewidth]{pic/mAP.pdf}}
%    \includegraphics[width=.5\linewidth]{pic/recall.pdf}\hfill
%    \includegraphics[width=.5\linewidth]{pic/mAP.pdf}
    \includegraphics[width=1.\linewidth]{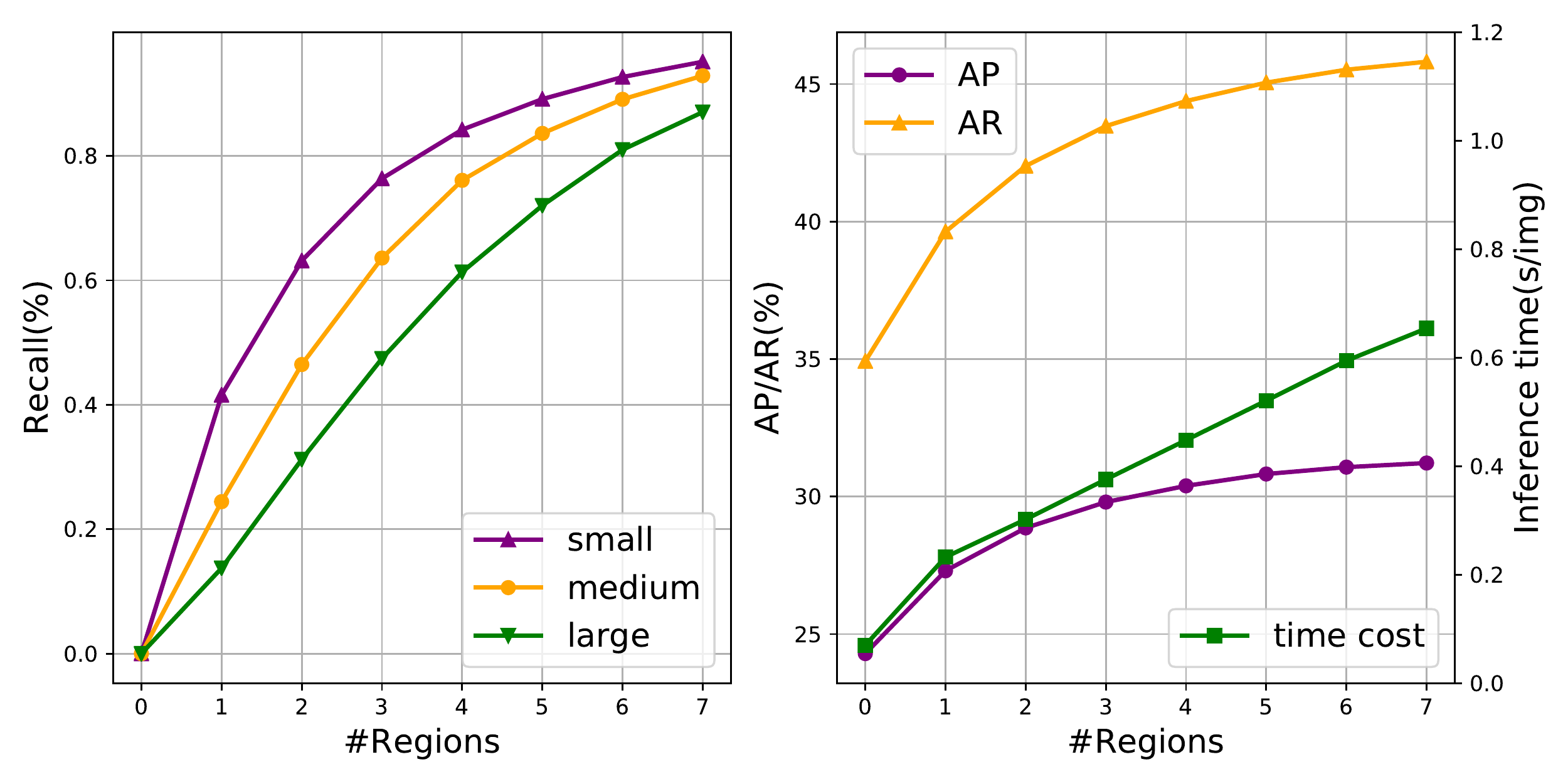}
\caption{\textbf{Left}: recall curves for objects \wrt the number of focus regions. \textbf{Right}: $AP/AR$ and inference time curves for the final detection performance on VisDrone2019 \textit{test-dev} dataset \wrt the number of focus regions. }
\label{recall}
\end{figure}

\begin{figure*}[ht]
  \centering
  \includegraphics[width=1.0\textwidth]{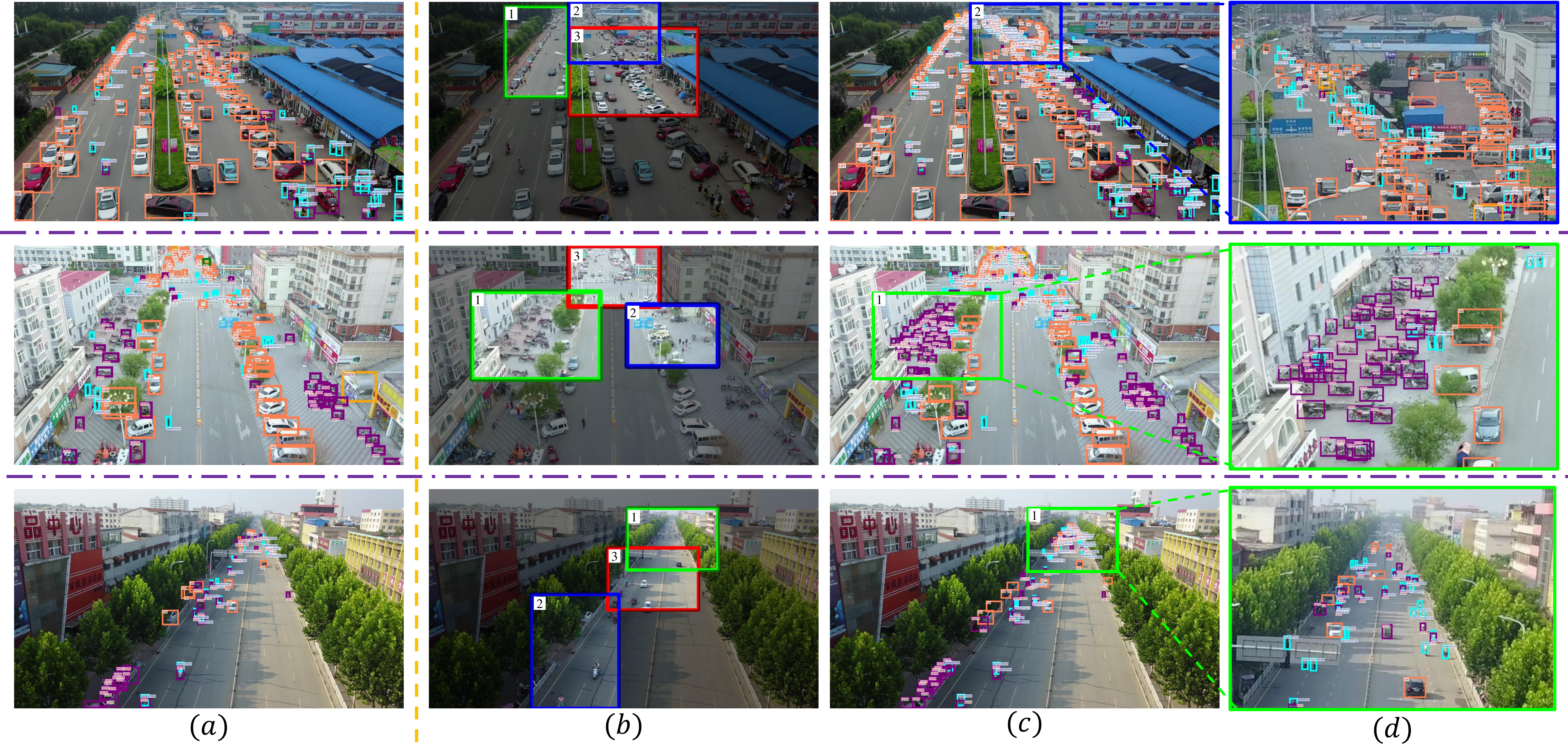}
  \caption{Visualizations of the performance of AdaZoom. (a) The detection results without AdaZoom. (b) The top-3 focus regions generated by AdaZoom. The focus regions adapt to the distribution and scales of objects. (c) The detection results with AdaZoom. The small and dense objects are also well detected. (d) The detection results on one of the focus regions. The results of (a) performs poorly on these regions.}\label{vis}
\end{figure*}

%\begin{table}[h]
%\begin{center}
%\footnotesize
%\setlength{\tabcolsep}{3.9pt}
%\begin{tabular}{|ccc|ccc|c|}
%  \hline
%    SA & RA &CT& $AP[\%]$ & $AP_{50}[\%]$  & $AP_{75}[\%]$  & $AR_{500}[\%]$  \\
%\hline\hline
%
%                               &   &    & 29.74 & 54.06  & 29.48 & 44.30 \\
%\hline
%                          \checkmark  &   &  & 29.90 & 54.37  & 29.65   & 44.45 \\
%                            & \checkmark  &  & 30.12 & 54.62  & 29.91  & 44.59 \\
%                            &   & \checkmark & 29.99 & 54.21  & 29.86   & 44.46 \\
%\hline
%                         \checkmark &\checkmark& & 30.51  & 55.28   & 30.22   & 45.15\\
%                         \checkmark &&\checkmark& 30.25  & 54.81   & 29.94  & 44.67\\
%                         &\checkmark& \checkmark & 30.56 & 55.04  & 30.64  & 44.81 \\
%\hline
%                          \checkmark & \checkmark&\checkmark& \textbf{31.22}  & \textbf{56.16}  & \textbf{31.22}  & \textbf{45.81} \\
%
%
%\hline
%\end{tabular}
%\end{center}
%\caption{Ablation for AdaZoom on VisDrone2019 \textit{test-dev}. We adopt Faster R-CNN with ResNet50 as detector. \textbf{SA}: scale adaptation of AdaZoom. \textbf{RA}: aspect ratio adaptation of AdaZoom. \textbf{CT}: collaborative training.}
%\label{ablation}
%\end{table}

\subsection{Ablation Study}

There are two main components for AdaZoom network, such as adaptive generation of focus regions and collaborative training between AdaZoom and the detector. We evaluate the effects of different components on VisDrone2019 \textit{test-dev} dataset. We use Faster R-CNN with ResNet50 for the ablation study, as shown in Table.\ref{ablation}. SR denotes focus regions with adaptive scales of $[240^2, 350^2, 420^2]$ and adaptive aspect ratios of $[0.7, 1.0, 1.5]$. CT denotes collaborative training. The baseline uses the single scale of $350^2$ and aspect ratio of 1.0 without collaborative training, achieving $29.74\%$ in $AP$.

\textbf{Effect of scale/ratio adaptation}. The scale and ratio adaptation of focus regions achieves the dynamic multi-scale detection. The scale adaptation alleviate the problem of scale variation. The adaptation of aspect ratio improves the recall of objects with the limited number of regions. As Table.\ref{ablation} shows, SR stably improves the $AP$. In particular, compared with baseline, SR increases the flexibility and diversity in region generation which improves $AP$ by $0.77\%$. Compared with the model with CT, SR further increased the $AP$ by $1.23\%$ and $AP_{50}$ by $1.95\%$. The results prove that SR improves the performance of detection by adaptively zooming the focus regions with flexible scales and aspect ratios. The detector benefits from adaptive-scale detection.

\textbf{Effect of collaborate training}. CT optimizes the detector with the focus regions generated by AdaZoom. Meanwhile, it guides AdaZoom to mine the fine-level regions for detector boosting with the reward based on detection results. Table.\ref{ablation} shows that CT gains consistent performance improvement under different settings. Compared to baseline, the CT slightly boosts the $AP$ by $0.25\%$ and $AP_{75}$ by $0.38\%$, because the scale of regions keeps the same. The collaborate training loses the benefits from multi-scale detection. When combining with SR, CT gains performance improvement as $0.71\%$ in $AP$ and $1.00\%$ in $AP_{75}$. The results show that involving collaboration between AdaZoom and the detector further promotes the object detection.

\textbf{Effect of number of focus regions.} The AdaZoom is motivated to adaptively propose focus regions with small and densely distributed objects, for further zoomed detection. To evaluate this, we present the recall rates (\ie, the proportion of objects that are enclosed in the generated regions) for objects of the reinforcement-learning-based AdaZoom, as shown in Fig.~\ref{recall} (left). For detailed analysis, we report the recall for \textit{small}, \textit{medium} and \textit{large} objects, respectively. As can be seen in Fig.~\ref{recall} (left), when the number of focus regions is 7, the recall of \textit{small}, \textit{medium} and \textit{large} achieve $95.1\%$, $92.9\%$, and $87.0\%$, respectively. Recall of small objects increasing fastest at the beginning. Because AdaZoom pays more attention to small objects and it prefers to focus on the regions with small objects.

With detector of ResNet50-based Faster R-CNN, the $AP$ and $AR$ are presented in Fig.~\ref{recall}(right). When the number of regions is zero, the detector is trained on the generated regions and infers on original images. The inference time almost linearly increases \wrt the number of focus regions. Our method can easily achieve the balance between accuracy and efficiency by setting the number of focus regions.

\textbf{Qualitative evaluation.} We visualize the detection results with and without AdaZoom in Fig.~\ref{vis}. We also display the top-3 focus regions for each image and the detection results on the focus regions. It can be observed from Fig.~\ref{vis} (b) that the AdaZoom generates small region for small objects and the shape of the region adapts to the distribution of objects. Comparing Fig.~\ref{vis} (c) with (a), the small and dense objects are well detected by our method, especially in the focus regions. Our method can adaptively focus on the important regions and achieve higher recalls of the small and dense objects.

\vspace{-0.1cm}
\section{Conclusion}
In this work, we propose an Adaptive Zoom (AdaZoom) network to zoom the focus regions with flexible scale and aspect ratio for multi-scale object detection. We optimize AdaZoom with policy gradient algorithm without additional annotations for focus regions. Moreover, we propose collaborative training to promote the coordination between AdaZoom and the detector which further improves the performance of detection. Without bells and whistles, our method achieves state-of-the-art on the VisDrone2019, UAVDT and DOTA datasets.

{\small
\bibliographystyle{ieee_fullname}
\bibliography{egbib}

\begin{thebibliography}{10}\itemsep=-1pt

\bibitem{bellman1957markovian}
Richard Bellman.
\newblock A markovian decision process.
\newblock {\em Journal of mathematics and mechanics}, pages 679--684, 1957.

\bibitem{2016Hierarchical}
Miriam Bellver, Xavier Giro-I-Nieto, Ferran Marques, and Jordi Torres.
\newblock Hierarchical object detection with deep reinforcement learning.
\newblock {\em Advances in Parallel Computing}, 31, 2016.

\bibitem{cai2016unified}
Zhaowei Cai, Quanfu Fan, Rogerio~S Feris, and Nuno Vasconcelos.
\newblock A unified multi-scale deep convolutional neural network for fast
  object detection.
\newblock In {\em European conference on computer vision}, pages 354--370.
  Springer, 2016.

\bibitem{cai2018cascade}
Zhaowei Cai and Nuno Vasconcelos.
\newblock Cascade r-cnn: Delving into high quality object detection.
\newblock In {\em Proceedings of the IEEE conference on computer vision and
  pattern recognition}, pages 6154--6162, 2018.

\bibitem{2015Active}
Juan~C. Caicedo and Svetlana Lazebnik.
\newblock Active object localization with deep reinforcement learning.
\newblock In {\em 2015 IEEE International Conference on Computer Vision
  (ICCV)}, 2015.

\bibitem{chen2019rrnet}
Changrui Chen, Yu Zhang, Qingxuan Lv, Shuo Wei, Xiaorui Wang, Xin Sun, and
  Junyu Dong.
\newblock Rrnet: A hybrid detector for object detection in drone-captured
  images.
\newblock In {\em Proceedings of the IEEE International Conference on Computer
  Vision Workshops}, pages 0--0, 2019.

\bibitem{dai2016r}
Jifeng Dai, Yi Li, Kaiming He, and Jian Sun.
\newblock R-fcn: Object detection via region-based fully convolutional
  networks.
\newblock In {\em Advances in neural information processing systems}, pages
  379--387, 2016.

\bibitem{deng2020global}
Sutao Deng, Shuai Li, Ke Xie, Wenfeng Song, Xiao Liao, Aimin Hao, and Hong Qin.
\newblock A global-local self-adaptive network for drone-view object detection.
\newblock {\em IEEE Transactions on Image Processing}, 2020.

\bibitem{du2018unmanned}
Dawei Du, Yuankai Qi, Hongyang Yu, Yifan Yang, Kaiwen Duan, Guorong Li, Weigang
  Zhang, Qingming Huang, and Qi Tian.
\newblock The unmanned aerial vehicle benchmark: Object detection and tracking.
\newblock In {\em Proceedings of the European Conference on Computer Vision
  (ECCV)}, pages 370--386, 2018.

\bibitem{gao2018dynamic}
Mingfei Gao, Ruichi Yu, Ang Li, Vlad~I Morariu, and Larry~S Davis.
\newblock Dynamic zoom-in network for fast object detection in large images.
\newblock In {\em Proceedings of the IEEE Conference on Computer Vision and
  Pattern Recognition}, pages 6926--6935, 2018.

\bibitem{2015Fast}
Ross Girshick.
\newblock Fast r-cnn.
\newblock {\em Computer ence}, 2015.

\bibitem{gonzalez2015active}
Abel Gonzalez-Garcia, Alexander Vezhnevets, and Vittorio Ferrari.
\newblock An active search strategy for efficient object class detection.
\newblock In {\em Proceedings of the IEEE Conference on Computer Vision and
  Pattern Recognition}, pages 3022--3031, 2015.

\bibitem{guo2020augfpn}
Chaoxu Guo, Bin Fan, Qian Zhang, Shiming Xiang, and Chunhong Pan.
\newblock Augfpn: Improving multi-scale feature learning for object detection.
\newblock In {\em Proceedings of the IEEE/CVF Conference on Computer Vision and
  Pattern Recognition}, pages 12595--12604, 2020.

\bibitem{hara2017attentional}
Kota Hara, Ming-Yu Liu, Oncel Tuzel, and Amir-massoud Farahmand.
\newblock Attentional network for visual object detection.
\newblock {\em arXiv preprint arXiv:1702.01478}, 2017.

\bibitem{2014Spatial}
Kaiming He, Xiangyu Zhang, Shaoqing Ren, and Jian Sun.
\newblock Spatial pyramid pooling in deep convolutional networks for visual
  recognition.
\newblock {\em IEEE Transactions on Pattern Analysis and Machine Intelligence},
  37(9):1904--16, 2014.

\bibitem{2016Deep}
Kaiming He, Xiangyu Zhang, Shaoqing Ren, and Jian Sun.
\newblock Deep residual learning for image recognition.
\newblock In {\em 2016 IEEE Conference on Computer Vision and Pattern
  Recognition (CVPR)}, 2016.

\bibitem{kong2017collaborative}
Xiangyu Kong, Bo Xin, Yizhou Wang, and Gang Hua.
\newblock Collaborative deep reinforcement learning for joint object search.
\newblock In {\em Proceedings of the IEEE Conference on Computer Vision and
  Pattern Recognition}, pages 1695--1704, 2017.

\bibitem{law2018cornernet}
Hei Law and Jia Deng.
\newblock Cornernet: Detecting objects as paired keypoints.
\newblock In {\em Proceedings of the European Conference on Computer Vision
  (ECCV)}, pages 734--750, 2018.

\bibitem{li2020density}
Changlin Li, Taojiannan Yang, Sijie Zhu, Chen Chen, and Shanyue Guan.
\newblock Density map guided object detection in aerial images.
\newblock In {\em Proceedings of the IEEE/CVF Conference on Computer Vision and
  Pattern Recognition Workshops}, pages 190--191, 2020.

\bibitem{li2018Object}
Yuezhang Li, Katia Sycara, and Rahul Iyer.
\newblock Object-sensitive deep reinforcement learning.
\newblock {\em arXiv preprint arXiv:1809.06064}, 2018.

\bibitem{lin2017feature}
Tsung-Yi Lin, Piotr Doll{\'a}r, Ross Girshick, Kaiming He, Bharath Hariharan,
  and Serge Belongie.
\newblock Feature pyramid networks for object detection.
\newblock In {\em Proceedings of the IEEE conference on computer vision and
  pattern recognition}, pages 2117--2125, 2017.

\bibitem{lin2014microsoft}
Tsung-Yi Lin, Michael Maire, Serge Belongie, James Hays, Pietro Perona, Deva
  Ramanan, Piotr Doll{\'a}r, and C~Lawrence Zitnick.
\newblock Microsoft coco: Common objects in context.
\newblock In {\em European conference on computer vision}, pages 740--755.
  Springer, 2014.

\bibitem{liu2018path}
Shu Liu, Lu Qi, Haifang Qin, Jianping Shi, and Jiaya Jia.
\newblock Path aggregation network for instance segmentation.
\newblock In {\em Proceedings of the IEEE conference on computer vision and
  pattern recognition}, pages 8759--8768, 2018.

\bibitem{liu2016ssd}
Wei Liu, Dragomir Anguelov, Dumitru Erhan, Christian Szegedy, Scott Reed,
  Cheng-Yang Fu, and Alexander~C Berg.
\newblock Ssd: Single shot multibox detector.
\newblock In {\em European conference on computer vision}, pages 21--37.
  Springer, 2016.

\bibitem{liu2020small}
Yingjie Liu, Fengbao Yang, and Peng Hu.
\newblock Small-object detection in uav-captured images via multi-branch
  parallel feature pyramid networks.
\newblock {\em IEEE Access}, 8:145740--145750, 2020.

\bibitem{lu2016adaptive}
Yongxi Lu, Tara Javidi, and Svetlana Lazebnik.
\newblock Adaptive object detection using adjacency and zoom prediction.
\newblock In {\em Proceedings of the IEEE Conference on Computer Vision and
  Pattern Recognition}, pages 2351--2359, 2016.

\bibitem{massa2018mrcnn}
Francisco Massa and Ross Girshick.
\newblock {maskrcnn-benchmark: Fast, modular reference implementation of
  Instance Segmentation and Object Detection algorithms in PyTorch}.
\newblock \url{https://github.com/facebookresearch/maskrcnn-benchmark}, 2018.
\newblock Accessed: [Insert date here].

\bibitem{Mathe_2016_CVPR}
Stefan Mathe, Aleksis Pirinen, and Cristian Sminchisescu.
\newblock Reinforcement learning for visual object detection.
\newblock In {\em Proceedings of the IEEE Conference on Computer Vision and
  Pattern Recognition (CVPR)}, June 2016.

\bibitem{mcconkie1975span}
George~W McConkie and Keith Rayner.
\newblock The span of the effective stimulus during a fixation in reading.
\newblock {\em Perception \& Psychophysics}, 17(6):578--586, 1975.

\bibitem{najibi2019autofocus}
Mahyar Najibi, Bharat Singh, and Larry~S Davis.
\newblock Autofocus: Efficient multi-scale inference.
\newblock In {\em Proceedings of the IEEE International Conference on Computer
  Vision}, pages 9745--9755, 2019.

\bibitem{pirinen2018deep}
Aleksis Pirinen and Cristian Sminchisescu.
\newblock Deep reinforcement learning of region proposal networks for object
  detection.
\newblock In {\em Proceedings of the IEEE Conference on Computer Vision and
  Pattern Recognition}, pages 6945--6954, 2018.

\bibitem{redmon2016you}
Joseph Redmon, Santosh Divvala, Ross Girshick, and Ali Farhadi.
\newblock You only look once: Unified, real-time object detection.
\newblock In {\em Proceedings of the IEEE conference on computer vision and
  pattern recognition}, pages 779--788, 2016.

\bibitem{ren2015faster}
Shaoqing Ren, Kaiming He, Ross Girshick, and Jian Sun.
\newblock Faster r-cnn: Towards real-time object detection with region proposal
  networks.
\newblock In {\em Advances in neural information processing systems}, pages
  91--99, 2015.

\bibitem{2017Object}
Shaoqing Ren, Kaiming He, Ross Girshick, Xiangyu Zhang, and Jian Sun.
\newblock Object detection networks on convolutional feature maps.
\newblock {\em IEEE Transactions on Pattern Analysis and Machine Intelligence},
  39(7):1476--1481, 2017.

\bibitem{singh2018analysis}
Bharat Singh and Larry~S Davis.
\newblock An analysis of scale invariance in object detection snip.
\newblock In {\em Proceedings of the IEEE conference on computer vision and
  pattern recognition}, pages 3578--3587, 2018.

\bibitem{singh2018sniper}
Bharat Singh, Mahyar Najibi, and Larry~S Davis.
\newblock Sniper: Efficient multi-scale training.
\newblock In {\em Advances in neural information processing systems}, pages
  9310--9320, 2018.

\bibitem{1999Policy}
Richard~S. Sutton, David Mcallester, Satinder Singh, and Yishay Mansour.
\newblock Policy gradient methods for reinforcement learning with function
  approximation.
\newblock {\em Submitted to Advances in Neural Information Processing Systems},
  12, 1999.

\bibitem{uzkent2020learning}
Burak Uzkent and Stefano Ermon.
\newblock Learning when and where to zoom with deep reinforcement learning.
\newblock In {\em Proceedings of the IEEE/CVF Conference on Computer Vision and
  Pattern Recognition}, pages 12345--12354, 2020.

\bibitem{wang2019spatial}
Haoran Wang, Zexin Wang, Meixia Jia, Aijin Li, Tuo Feng, Wenhua Zhang, and
  Licheng Jiao.
\newblock Spatial attention for multi-scale feature refinement for object
  detection.
\newblock In {\em Proceedings of the IEEE International Conference on Computer
  Vision Workshops}, pages 0--0, 2019.

\bibitem{wang2020object}
Yi Wang, Youlong Yang, and Xi Zhao.
\newblock Object detection using clustering algorithm adaptive searching
  regions in aerial images.
\newblock In {\em European Conference on Computer Vision}, pages 651--664.
  Springer, 2020.

\bibitem{xia2018dota}
Gui-Song Xia, Xiang Bai, Jian Ding, Zhen Zhu, Serge Belongie, Jiebo Luo, Mihai
  Datcu, Marcello Pelillo, and Liangpei Zhang.
\newblock Dota: A large-scale dataset for object detection in aerial images.
\newblock In {\em Proceedings of the IEEE Conference on Computer Vision and
  Pattern Recognition}, pages 3974--3983, 2018.

\bibitem{yang2019clustered}
Fan Yang, Heng Fan, Peng Chu, Erik Blasch, and Haibin Ling.
\newblock Clustered object detection in aerial images.
\newblock In {\em Proceedings of the IEEE International Conference on Computer
  Vision}, pages 8311--8320, 2019.

\bibitem{2016AttentionNet}
Donggeun Yoo, Sunggyun Park, Joon~Young Lee, Anthony~S Paek, and In~So Kweon.
\newblock Attentionnet: Aggregating weak directions for accurate object
  detection.
\newblock In {\em 2015 IEEE International Conference on Computer Vision
  (ICCV)}, 2016.

\bibitem{zhang2019fully}
Junyi Zhang, Junying Huang, Xuankun Chen, and Dongyu Zhang.
\newblock How to fully exploit the abilities of aerial image detectors.
\newblock In {\em Proceedings of the IEEE International Conference on Computer
  Vision Workshops}, pages 0--0, 2019.

\bibitem{zhang2019dense}
Xindi Zhang, Ebroul Izquierdo, and Krishna Chandramouli.
\newblock Dense and small object detection in uav vision based on cascade
  network.
\newblock In {\em Proceedings of the IEEE International Conference on Computer
  Vision Workshops}, pages 0--0, 2019.

\bibitem{zhou2019scale}
Jingkai Zhou, Chi-Man Vong, Qiong Liu, and Zhenyu Wang.
\newblock Scale adaptive image cropping for uav object detection.
\newblock {\em Neurocomputing}, 366:305--313, 2019.

\bibitem{zhu2018vision}
Pengfei Zhu, Longyin Wen, Xiao Bian, Haibin Ling, and Qinghua Hu.
\newblock Vision meets drones: A challenge.
\newblock {\em arXiv preprint arXiv:1804.07437}, 2018.

\end{thebibliography}
}

\end{document}